\documentclass[conference]{IEEEtran}
\usepackage{blindtext, graphicx}
\usepackage{amssymb} 
\usepackage[backend=bibtex,style=ieee,sorting=ynt]{biblatex}
\usepackage{color}
\newcommand{\todo}[1]{}
\renewcommand{\todo}[1]{{\color{red} TODO: {#1}}}

\def\LRT#1#2{\!
\raisebox{.2ex}{$
{{\scriptstyle\;#1}\atop{\displaystyle\gtrless}}
\atop
{\raisebox{-1.25ex}{$\scriptstyle\;#2$}}
$}
\!}

\begin{document}
\title{Recurrent Neural Radio Anomaly Detection}

\author{\IEEEauthorblockN{Timothy J. O'Shea}
\IEEEauthorblockA{Bradley Department of Electrical \\ 
and Computer Engineering\\
Virginia Tech, Arlington, VA\\
Email: oshea@vt.edu}
\and
\IEEEauthorblockN{T. Charles Clancy}
\IEEEauthorblockA{Bradley Department of Electrical \\ 
and Computer Engineering\\
Virginia Tech, Arlington, VA\\
Email: tcc@vt.edu}
\and
\IEEEauthorblockN{Robert W. McGwier}
\IEEEauthorblockA{Bradley Department of Electrical \\ 
and Computer Engineering\\
Virginia Tech, Arlington, VA\\
Email: rwmcgwi@vt.edu}
}

\maketitle

\begin{abstract}
We introduce a powerful recurrent neural network based method for novelty detection to the application of detecting radio anomalies.  This approach holds promise in significantly increasing the ability of naive anomaly detection to detect small anomalies in highly complex complexity multi-user radio bands.  We demonstrate the efficacy of this approach on a number of common real over the air radio communications bands of interest and quantify detection performance in terms of probability of detection an false alarm rates across a range of interference to band power ratios and compare to baseline methods.
\end{abstract}

\IEEEpeerreviewmaketitle

\section{Introduction}
Anomaly detection is an important time-series function which is widely used in network security monitoring, medical sensor monitoring, financial change modeling, and any number of other applications.  There is a significant body of existing work on the subject in theory and applications \cite{marsland2003novelty} \cite{pimentel2014review}, we focus here primarily on reconstruction based novelty detection.  In radio, applications of anomaly detection have been discussed \cite{mitola2006cognitive} but not widely used outside of a few small niche applications.  

Radio anomaly detection has been leveraged somewhat in wireless sensor networks such as in \cite{2011RFanomalysurvey} \cite{patcha2007overview} \cite{zhang2000intrusion} \cite{rajasegarar2008anomaly}, but most of these applications focus on detecting changes in sensor data (temperature, pressure, etc), or expert features rather than on anomalies occurring in the high rate raw physical layer radio signal itself.  We are unaware of any currently widely used or investigated applications of naive anomaly detection within the raw unmodified radio physical layer rather than on an expert feature such as a detection statistic.  The focus on raw RF rather than a specialized statistic is based on the hope that such a technique may be able to generalize well to numerous types of signals, rather than relying on specific analysis to help a single scenario.

There are a number of driving motivations for such a capability; for instance for spectrum access enforcement by regulatory bodies in which the appearance of new radio users of any kind of emission on unauthorized bands or with unauthorized equipment or techniques presents an enforcement event which should be rapidly addressed.  In commercial and defense communications applications radio anomaly detection also presents the opportunity to rapidly recognize interfering emitters, malfunctioning equipment, or malicious users within their licensed bands and take action.  Each of these applications currently requires expensive high maintenance expert systems which often perform a series of steps of energy detection, localization, classification, and comparison to baseline databases, and alerting with large amounts of specialization and tuning to the band and signals of interest and taking significant computational horsepower and implementation expense to deploy.  By shifting these systems to more generally applicable naive methods using neural networks which can be highly optimized on concurrent architectures such as graphics processing units (GPUs) in a very general way, we offer the potential to make such systems much easier to realize, adapt to new domains, and run leveraging economies of scale on computing power and model primitive optimization.

\section{Approach}
\begin{figure}[!ht]
\centering
\includegraphics[width=0.4\textwidth]{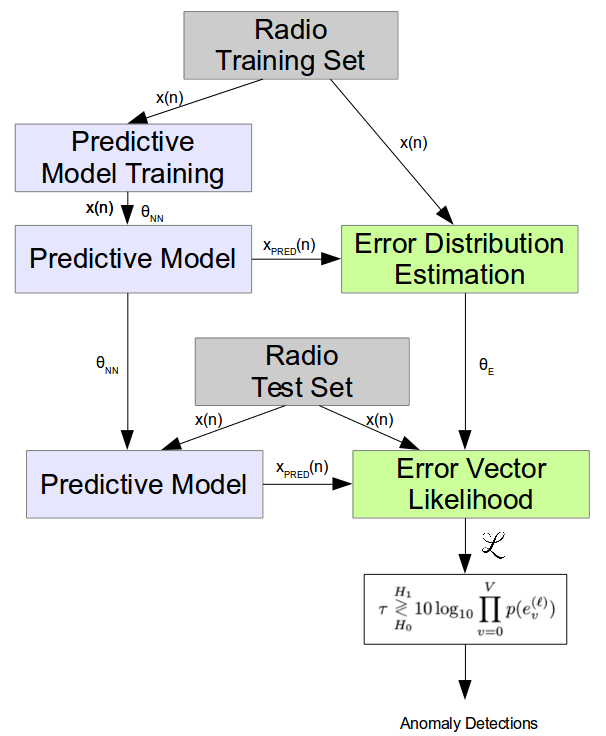}
\caption{Neural Radio Anomaly Detection System Model}
\label{fig:system}
\end{figure}

Recently, the use of recurrent neural networks to form a predictive reconstruction as part of a novelty detector has been proposed \cite{lstmad} and demonstrated to function quite well on several time series datasets including electrocardiograms, physical telemetry signals, and power consumption metrics.  In this work a long short term memory \cite{lstm} (LSTM) based recurrent network is used to train a time series predictor on a training set which is then used to compute an expected error distribution vs the real signal which is well characterized for non-anomalous behavior.  The sequence learning capacity of LSTM has in some cases been shown to exceed that of Hidden Markov model \cite{graves2009novel} and Kalman based linear predictors, because it is able to take into account a much more complex nonlinear representation of state (does not make the Markov assumption), short and long term transition dependencies, and complex nonlinear output mapping dependencies than either of these prior models are capable of.  An overview of this system level model is shown in figure \ref{fig:system}.

In this case we consider a sampled radio data time series $X = \{x_1, x_2, ..., x_M\}$, our training support, where each point is a complex base-band sample in $\mathbb{R}^2$.  We train a predictor $f$ with learned parameters $\theta$ such that for any start offset $k$, input samples count $N$, and prediction length $\ell$, we form a sequence regression problem shown in equation \ref{eq:0}.

\begin{equation} \label{eq:0}
\{\hat{x}_{k+N}, ..., \hat{x}_{k+N+\ell-1} \} = f\left( \{x_{k}, ..., x_{k+N-1}, \theta_{NN} \right) 
\end{equation}

This function $f$ can take the form of any number of different prediction methods, but in this case we evaluate a recurrent neural network (RNN) based predictor leveraging LSTM layers.

A predictor error vector can then be obtained on the training set by computing the difference $e_i = x_i - \hat{x}_i $ for each predicted value given the known (non-causal) actual values of $x$ over $[k+N,k+N+\ell-1]$.  This error vector is used to estimate of distribution $\{e_{0},...,e_{\ell-1} \} \sim P_e(e^{(\ell)}, \theta_E)$ through some form of parametric or non-parametric density estimation.  In this case, we model using a parametric multivariate Gaussian distribution.

After fitting both the predictive model parameters $\theta_NN$ and the error distribution parameters $\theta_E$ on the dataset, we can now use this model to perform novelty detection by observing regions within the signal $x(n)$ where predictor error deviates significantly from the expected error distribution $D_e \triangleq p(e^{(\ell)}, \theta_E)$.  That is to say, we wish to compute $p(x_i \sim D_e)$.  

This can be done by thresholding the log-likelihood of $e_i$ in $D_e$ at some level to make a decision, or by combining $V$ sequential samples of the likelihood and thresholding the aggregate statistic.  Our expression used looks roughly like that given below in equation \ref{eq:1}.

\begin{equation} \label{eq:1}
    \tau \LRT{H_1}{H_0} 10 \log_{10} \prod_{v=0}^{V} p(e_v^{(\ell)})
\end{equation}

Here $H_1$ is the hypothesis that the current values of ${x_i,...,x_{i+N}}$ for each of the $V$ observations are drawn from a distribution matching that of our training set, representing 'normal' behavior, while $H_0$ is the hypothesis that the current values of ${x_i,...,x{i+N} }$ are not drawn from the distribution of the training set, representing an anomaly or novelty being present.  Our threshold $\tau$ can be fit using an $F_\beta$ or typical constant false alarm rate (CFAR) sorts of analysis \cite{hansen1973constant} if a decent dataset of "anomalous" behavior is known, or a false alarm rate on "normal" behavior is used as the metric.

\section{Wide-band Radio Communications Time-series}

Time-series in the radio domain are quite complex models, especially when considering wideband aggregates of numerous channels on separate frequencies, each with their own temporal and spectral channel access scheme, and each carrying whitened randomized data-bits which typically do not repeat aside from reference tones such as preamble, low-entropy-headers and pilots.  A time sequence prediction model for such an aggregate signal must then be able to account for short-time expected symbol transitions and pulse shaping of a each carrier and its channel variations as well as the symbols forming higher level traffic and application sequences representing behavior of users.  At both layers we must be able to model the sum of all users and emitters combined into a single shared medium on one or more channels.  Differing levels of predictor complexity and predictive capacity define how many of these effects of 'normal' behavior are modeled, defining the scale, complexity, or distance from the norm of the anomaly which can be detected using that model.

\begin{figure}[!ht]
\centering
\includegraphics[width=0.5\textwidth]{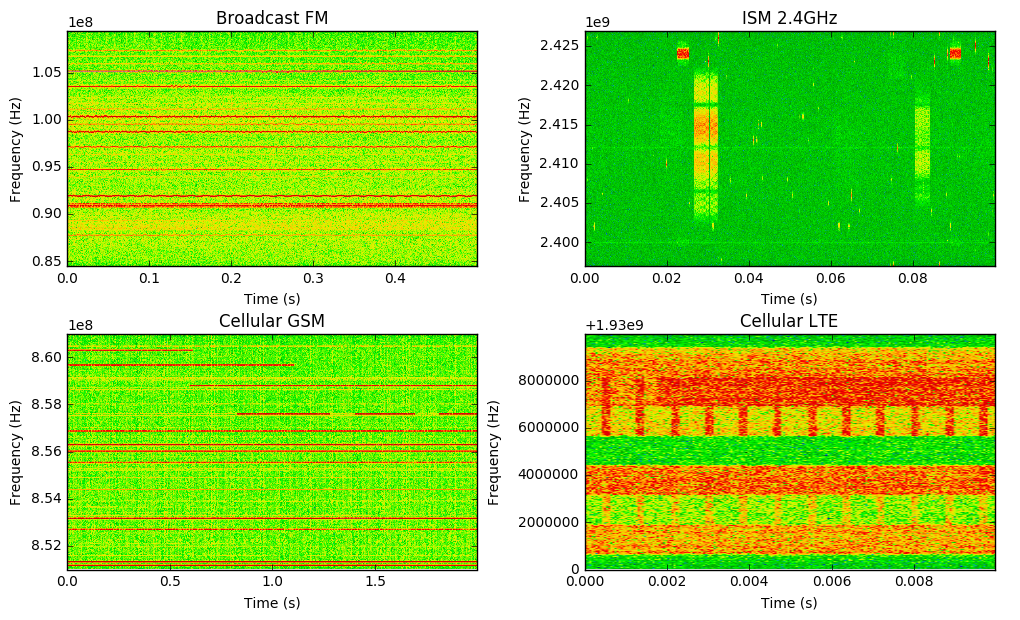}
\caption{Spectrogram plots of excerpts from radio example sequences}
\label{fig:rfseq}
\end{figure}

In figure \ref{fig:rfseq} we show spectrogram plots demonstrating a variety of radio time-series complexities of real world signals which we further consider in thi paper.  Here we see examples ranging from the most static, continuously modulated analog FM broadcast carriers, through relatively well structured on a macro-scale but extremely complexly coded and changing on a micro-scale, cellular band carriers of both GSM and LTE with its rapidly changing resource block (RBs) allocations in OFDMA time-frequency slots, and finally to the most chaotic ISM band environment comprised by CSMA/CD WiFi/IEEE802.11 bursts occurring at random times, and frequency hopped BlueTooth/IEEE802.15.1 bursts occurring at random times and frequencies among other emitters. 

Each of these bands presents different complexities which a temporal sequence prediction model needs to capture to  accurately form a predictive model for signal behavior.  We will consider each of these examples while evaluating our anomaly detection performance and train using these real recorded over the air RF datasets including harsh urban channel fading conditions.  Recordings are conducted with an Ettus Research B200-mini \cite{usrp} which uses the Analog Devices AD9361 RFIC front-end and stored to disk for analysis using GNU Radio \cite{gnuradio}.

\section{Predictor Models}

Here we describe each model $f\left( \{x_{k}, ..., x_{k+N-1}, \theta_{NN} \right)$ which we use to predict our next samples of the time-series sequence.  For fairness of comparison we normalize each predictor to 32 samples of input and 4 predicted output samples.  We include a several baseline models as well as a number of models modeled after state of the art time series learning neural network capabilities.  

\subsection{Kalman Sequence Predictor}

We use a 3rd order Unscented Kalman Filter/Predictor similar to that described in \cite{vittaldev2012second}.  This is implemented using the FilterPy module \cite{filterpy} and forms our performance benchmark for this paper.  This implements a traditional Kalman Novelty Detector as one might do without a learned predictive model.  In this case the adaptive filter is tuned online while running and the error distribution is characterized on this.

\subsection{DNN Sequence Predictor}

\begin{figure}[!ht]
\centering
\includegraphics[width=0.4\textwidth]{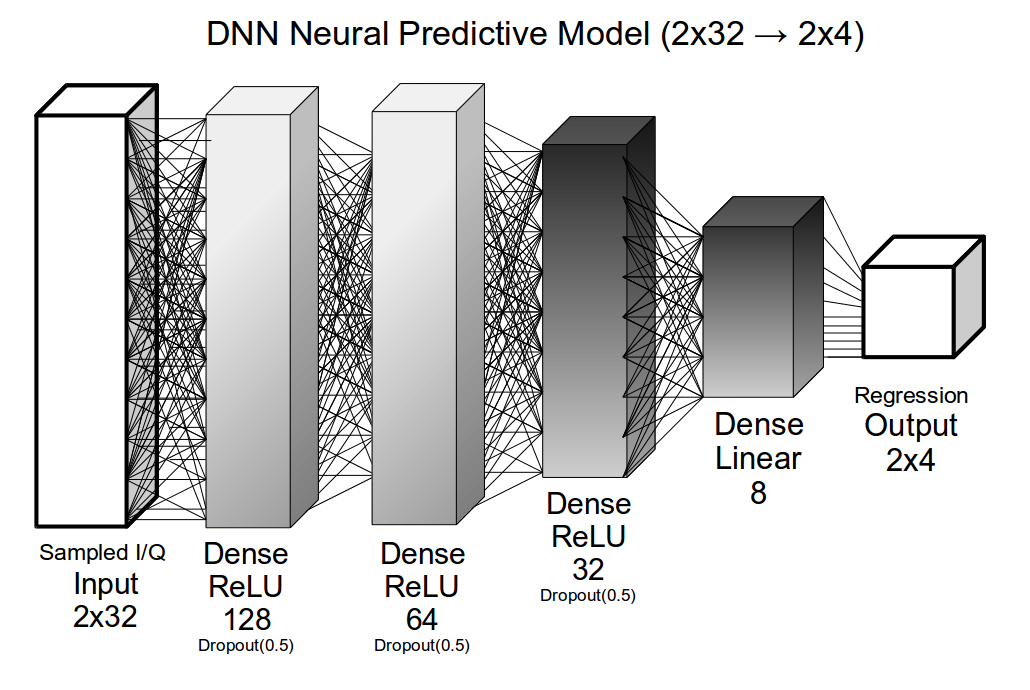}
\caption{DNN Predictor Network Architecture}
\label{fig:dnnnet}
\end{figure}

In our Dense Neural Network (DNN) model (shown in figure \ref{fig:dnnnet}), we train a naive fully-connected network as a neural network baseline with a high number of free parameters and heavy dropout allowing it to learn a completely unconstrained mapping between input and output samples.  This will allow us to compare other specialized/constrained architectures such as convolutional and recurrent varieties for model fitting appropriateness.

\subsection{Raw LSTM Sequence Predictor}

\begin{figure}[!ht]
\centering
\includegraphics[width=0.4\textwidth]{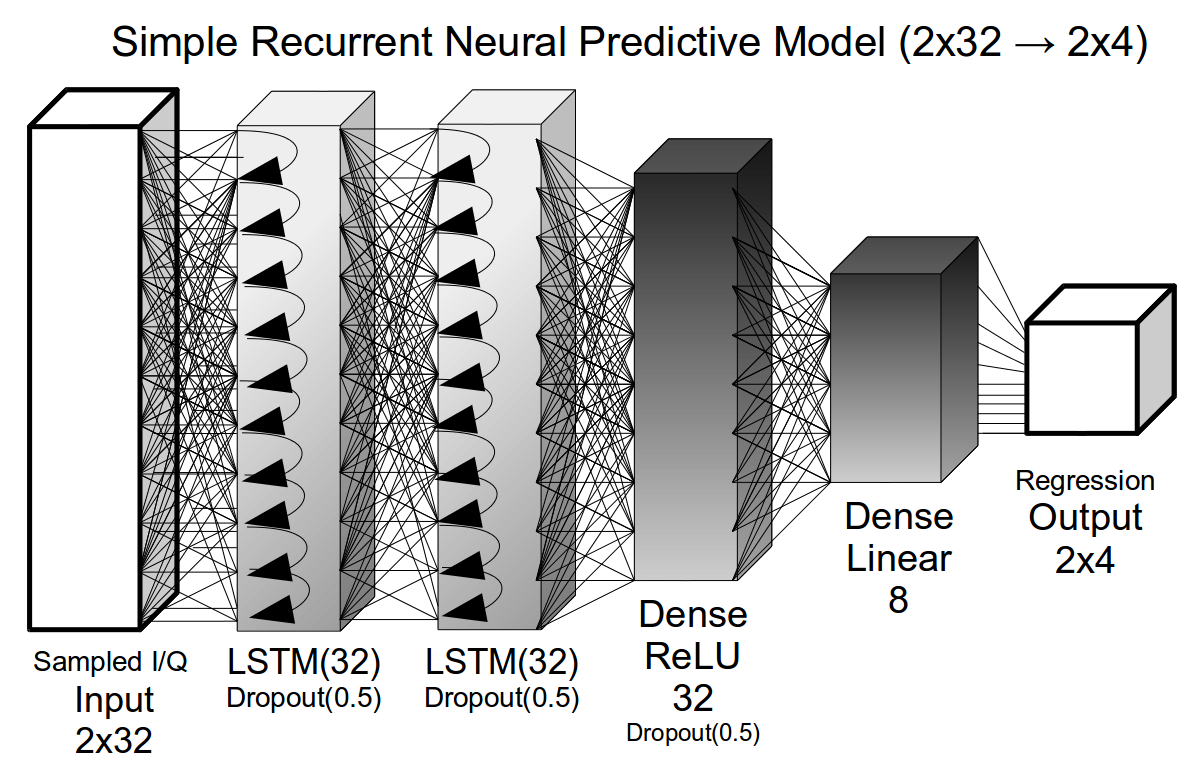}
\caption{LSTMArchitecture used for evaluation recurrent neural prediction network}
\label{fig:lstmnt}
\end{figure}

In the LSTM based sequence predictor model (Figure \ref{fig:lstmnt}), we implement a 2-layer LSTM followed by 2 fully connected layers culminating in a linear activation for regression of complex continuous valued sample output values.  We regularize between each layer with dropout of 0.5 and using proper LSTM weight and activation dropout as described in \cite{zaremba2014recurrent} and implemented in Keras \cite{keras}.

\subsection{DCNN1 Sequence Predictor}

\begin{figure}[!ht]
\centering
\includegraphics[width=0.4\textwidth]{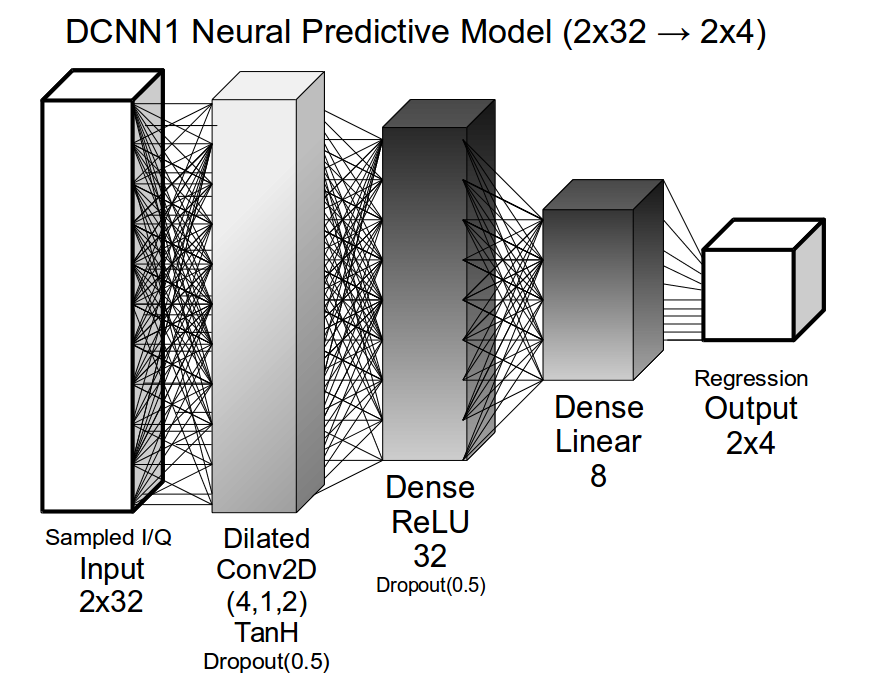}
\caption{DCNN1 Predictor Network Architecture}
\label{fig:dcnn1net}
\end{figure}

In the Dilated Convolutional Neural Network 1 (DCNN1) (Figure \ref{fig:dcnn1net}), we introduce a simple dilated convolution layer on the front end of a simple fully connected neural network to allow for learning of convolutional features at a stride of 2.

\subsection{DCNN2 Sequence Predictor}

\begin{figure}[!ht]
\centering
\includegraphics[width=0.4\textwidth]{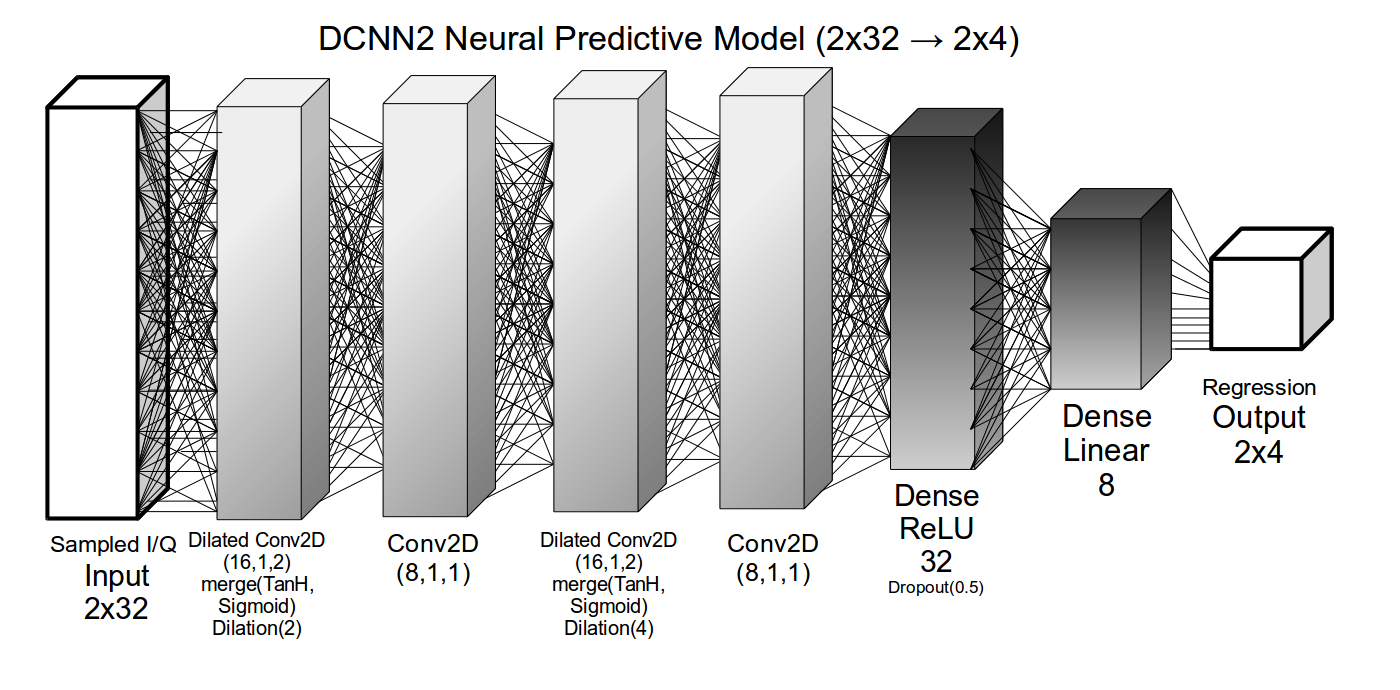}
\caption{DCNN2 Predictor Network Architecture}
\label{fig:anomnet}
\end{figure}

We model our Dilated Convolutional Neural Network 2 (DCNN2) architecture on a vastly simplified version of Google's WaveNet architecture \cite{wavenet} which has demonstrated a strong ability to learn raw time series representations on acoustic voice data.  Here we use two levels of dilated convolutions where each is a residual block \cite{resnets} containing identical layers with hyperbolic tan (TanH) and sigmoid activations merged multiplicatively, followed by a 1x1 convolutional layer for dimensionality reduction.  

\section{Model Optimization}

Before evaluating detection performance, we simply attempt to minimize the mean squared error of the predictor function to select our network parameters $\theta_{NN}$ and our architecture and hyper-parameters for the predictor.  From initial experimentation, optimal network architectures seem to vary slightly from dataset to dataset, but for now we seek to use a single set of network parameters for all datasets.

A much more extensive hyper-parameter search is really desired here to find best suited network structures.  We hope in future work to do this more extensively, within the scope of this work we only try a handful of architectures derived from proven architectures in prior work on similar tasks.

\section{Performance Evaluation}

To evaluate performance, we introduce a number of different classes of synthetic anomalies into each recorded sampled RF dataset and measure detection and false alarm rates using various methods of novelty detection.  Our anomaly types considered here span the range of time-frequency support from an instantaneous wide-band pulse, to a narrow-band tone at a single frequency, but are each normalized by total power of the same time support over the window $[t_s,t_e)$.  The anomaly classes considered are:

\begin{itemize}
    \item Pulsed Complex Sinusoid: expressed as $n(t) = exp(j 2 \pi t F_c/F_s)$ for $t \in [t_s, t_e)$ where $F_c \sim Uniform(-F_s/2,F_s/2)$.
    \item Short-time Broadband Bursts (Sinc pulse): expressed as $n(t) = sinc(2 \pi (t-(t_s+t_e)/2) Fc/Fs)$ for $t \in [t_s, t_e)$.
    \item Brief Periods of Signal Non-Linear Compression: approximated as $n(t) = 13x(t)-3x^3(t)$ for $t \in [t_s, t_e)$.
    \item Pulsed QPSK Signals: where $symrate \sim Uniform(F_s/250,F_s/2)$, $F_c \sim Uniform(-(F_s-symrate/2)/2, (F_s-symrate/2)/2)$, and a root-raised cosine pulse shaping filter of $\alpha=0.3$ and $N=11$ is applied at the baudrate.
    \item Pulsed Chirp Events: $n(t) = exp(j 2 \pi t F_c/F_s)$ for $t \in [t_s, t_e)$ where $F_c$ varies linearly in time from $F_{c1} \sim Uniform(-F_s/2,F_s/2)$ to $F_{c2} \sim Uniform(-F_s/2,F_s/2)$
\end{itemize}

\begin{figure}[!ht]
\centering
\includegraphics[width=0.4\textwidth]{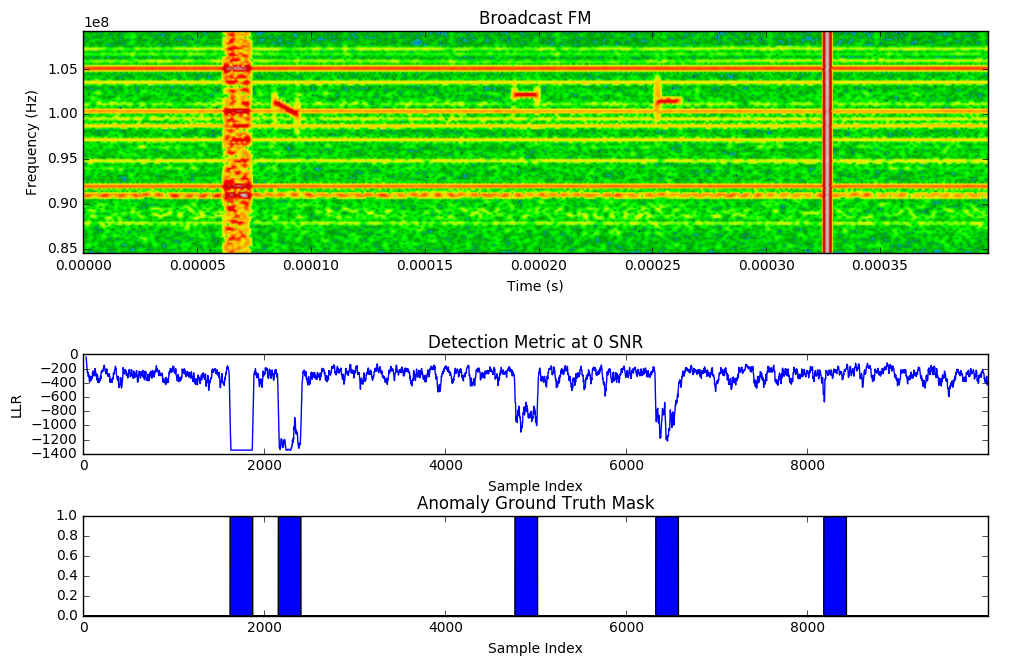}
\caption{Example synthetic anomalies on the FM broadband dataset, from left to right: compression, chirp, tone, qpsk, pulse}
\label{fig:anom1}
\end{figure}

We characterize each of these anomalies by its interference-to-band-power ratio (IBR) in dB.  We refer to this in some instances as signal-to-noise ratio (SNR) for convenience but the signal of interest here is the anomaly and the "noise" in this case is the power of the non-anomalous band including all signals therein.

Inspecting figure \ref{fig:anom1} we can see that performance does vary based on the anomaly type.  For instance, performance on non-linear compression, chirp detection, tone detection, and QPSK burst detection, all appear to be quite a bit stronger in the LLR detection metric than the wideband pulsed noise which is very short in time and results in an anomaly spike which is likewise extremely short in time.  We include several runs of bands below in figures \ref{fig:fmanom}, \ref{fig:lteanom}, and \ref{fig:gsmanom} for visual inspection.

\begin{figure}[!ht]
\centering
\includegraphics[width=0.4\textwidth]{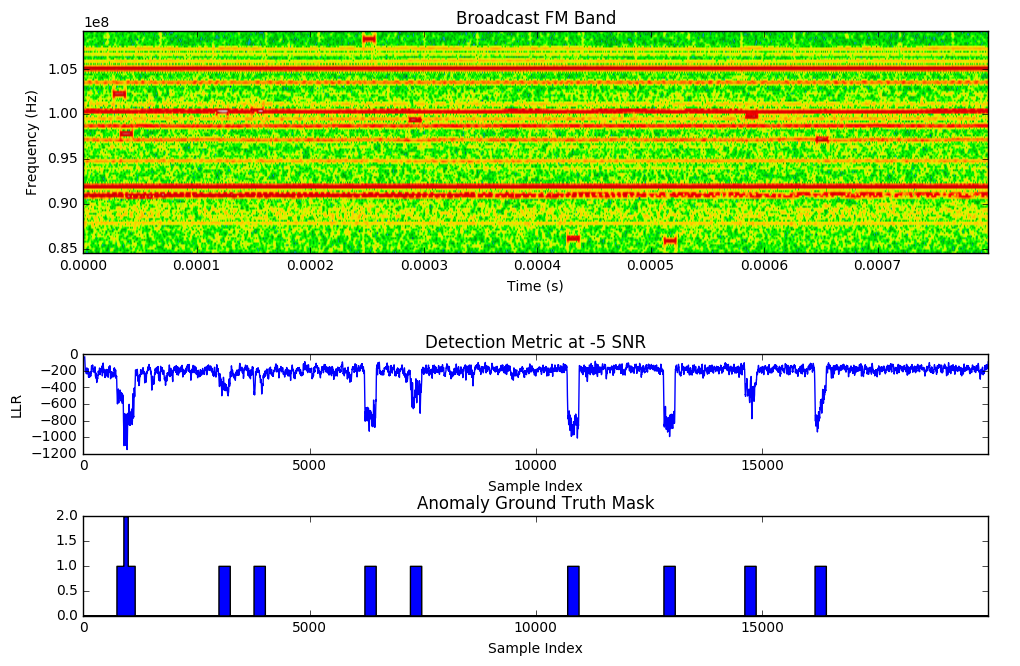}
\caption{LSTM Anomaly Detector on FM Band}
\label{fig:fmanom}
\end{figure}

\begin{figure}[!ht]
\centering
\includegraphics[width=0.4\textwidth]{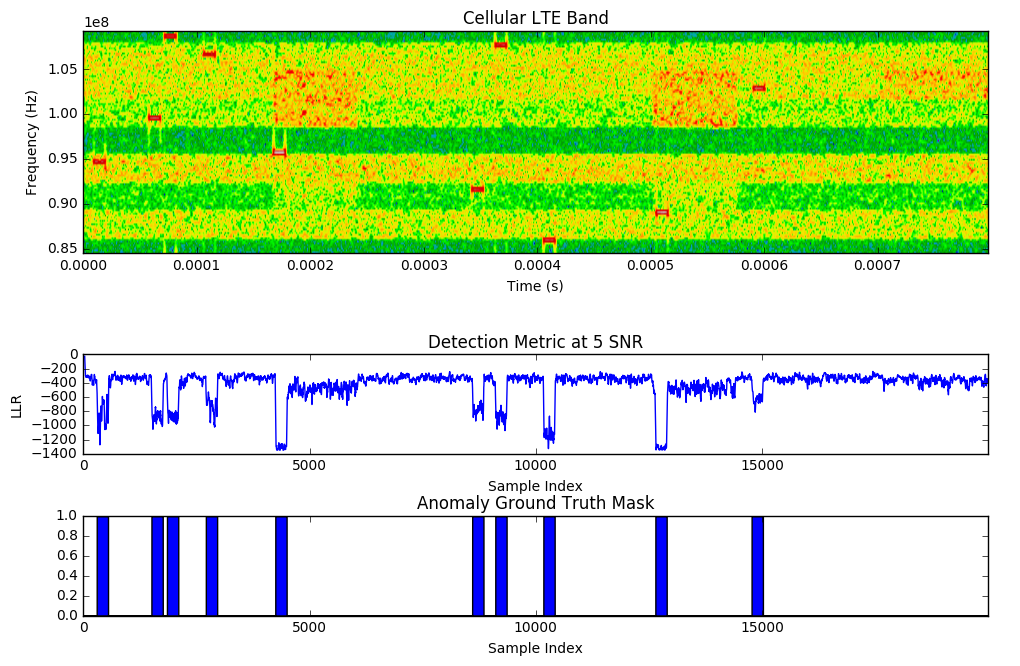}
\caption{LSTM Anomaly Detector on LTE Band}
\label{fig:lteanom}
\end{figure}

\begin{figure}[!ht]
\centering
\includegraphics[width=0.4\textwidth]{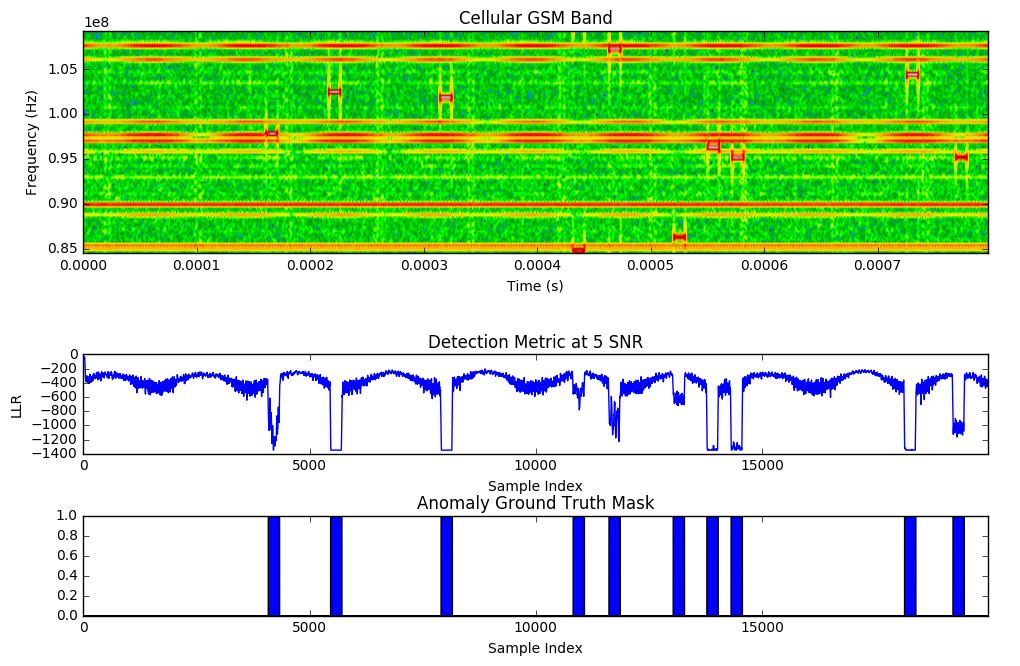}
\caption{LSTM Anomaly Detector on GSM Band}
\label{fig:gsmanom}
\end{figure}

\begin{figure}[!ht]
\centering
\includegraphics[width=0.4\textwidth]{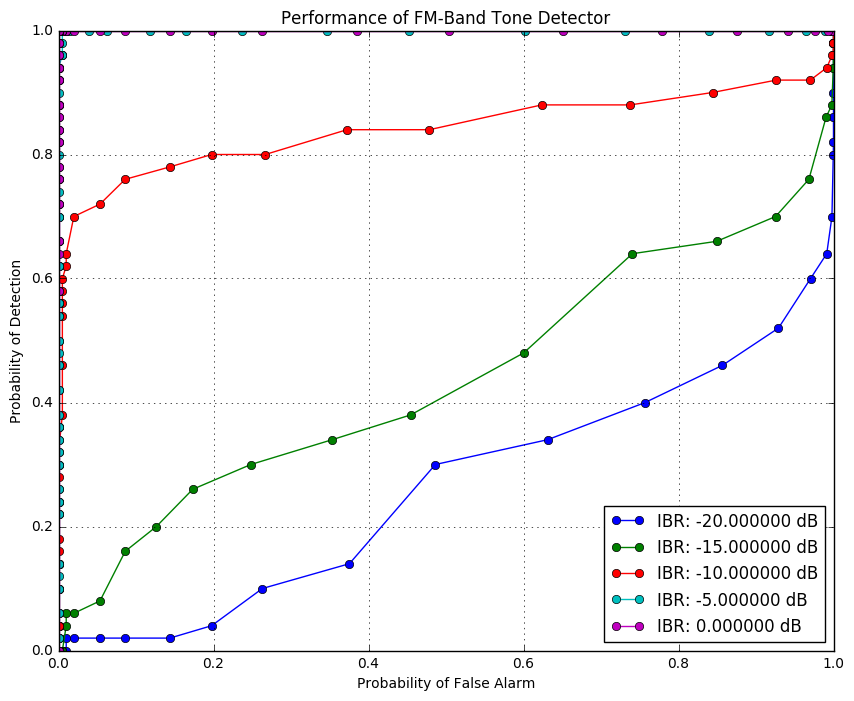}
\caption{Probability of Detection and False Alarm for sinusoidal tones on FM Broadcast Band}
\label{fig:pdpfa1}
\end{figure}

\begin{figure}[!ht]
\centering
\includegraphics[width=0.5\textwidth]{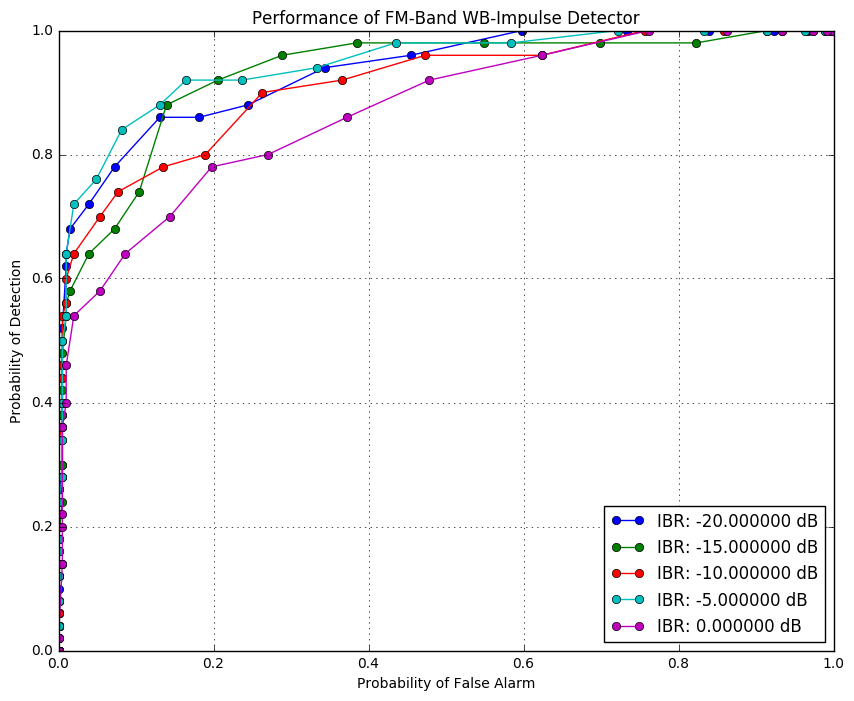}
\caption{Probability of Detection and False Alarm for wide-band pulses on FM Broadcast Band}
\label{fig:pdpfa2}
\end{figure}

In figure \ref{fig:pdpfa1} we show the performance of the tone detector across 100,000 samples of FM recording inserting 50 random tone events of length 250 samples.  As the IBR approaches -5dB, we have nearly perfect Pd/Pfa performance, while in figure \ref{fig:pdpfa2}, we see for the wideband pulse tone, which has very large instantaneous peak power but a very narrow time-support, the IBR does not have nearly as significant an impact on Pd/Pfa performance at these IBR levels.  In this case, our probability of detection represents the probability of detecting all anomalies present in the time range, while our probability of false alarm represents the probability of a false alarm being triggered in any 250-sample window.

\begin{figure}[!ht]
\centering
\includegraphics[width=0.5\textwidth]{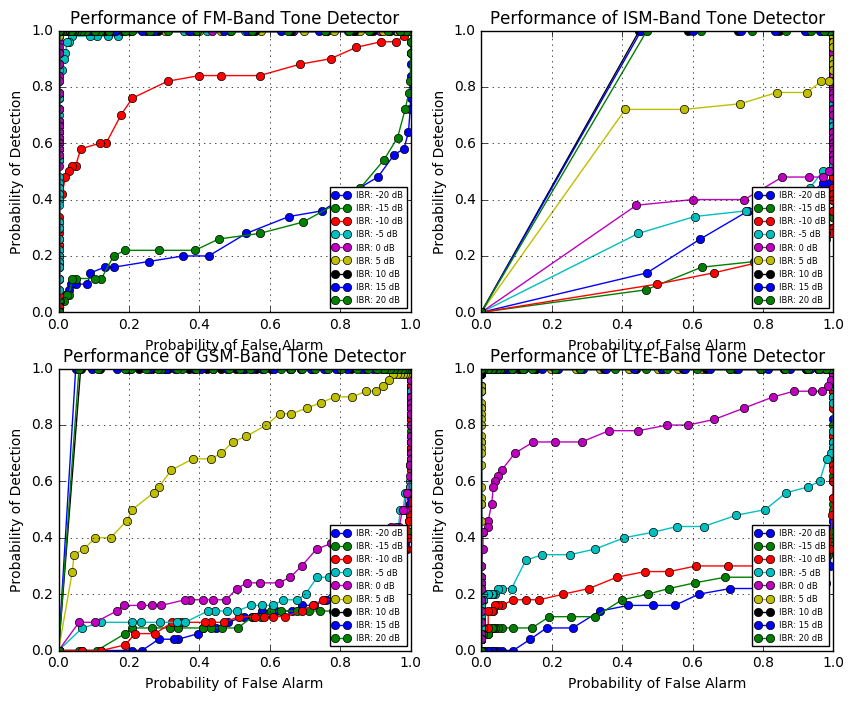}
\caption{Probability of Detection and False Alarm for Tones in each band type}
\label{fig:pdpfa3}
\end{figure}

We can repeat these experiments across a range of interference-to-band power ratios to observe the efficacy in a range of different modulation and multi-access schemes.  Results for this are shown below in figure \ref{fig:pdpfa3}.  Here we can see that for all channel types are relatively effective once we approach 0-5dB IBR.  The most difficult here is the ISM band, where our predictive model is used to seeing bursty CSMA/CD kinds of traffic from WiFi and blue-tooth frequency hopped bursts across the band.  In this case our anomaly detection ability is the most challenged of all the other bands.

\begin{figure}[!ht]
\centering
\includegraphics[width=0.5\textwidth]{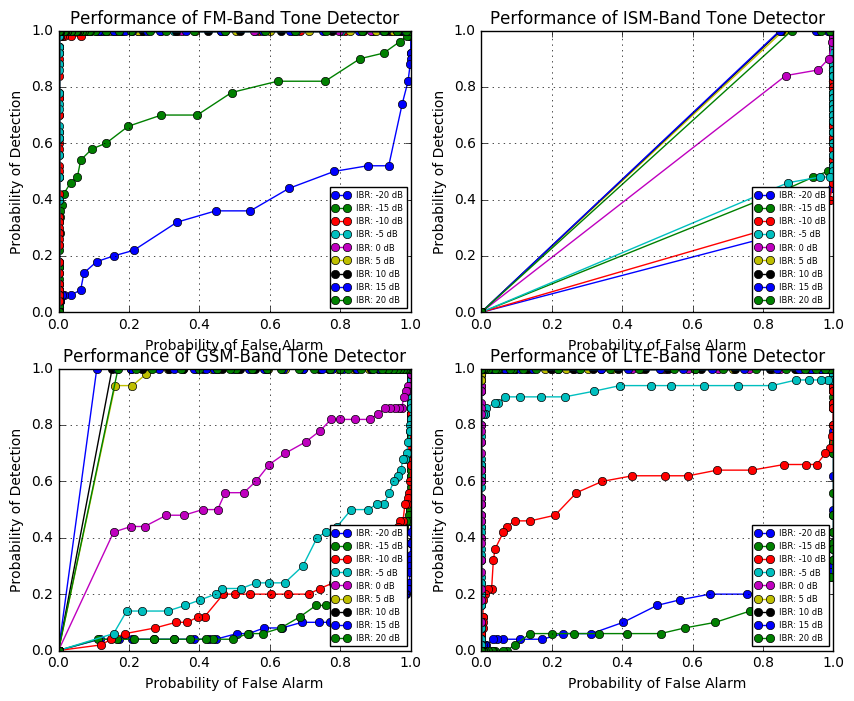}
\caption{Probability of Detection and False Alarm for Chirps in each band type}
\label{fig:pdpfa4}
\end{figure}

Repeating this experiment with chirp interference instead of pulse interference, we show performance in figure \ref{fig:pdpfa4}.  Again we see excellent performance above 0-5dB in most cases, although the ISM band continues to be the most difficult.

\begin{figure*}[!ht]
\centering
\includegraphics[width=0.75\textwidth]{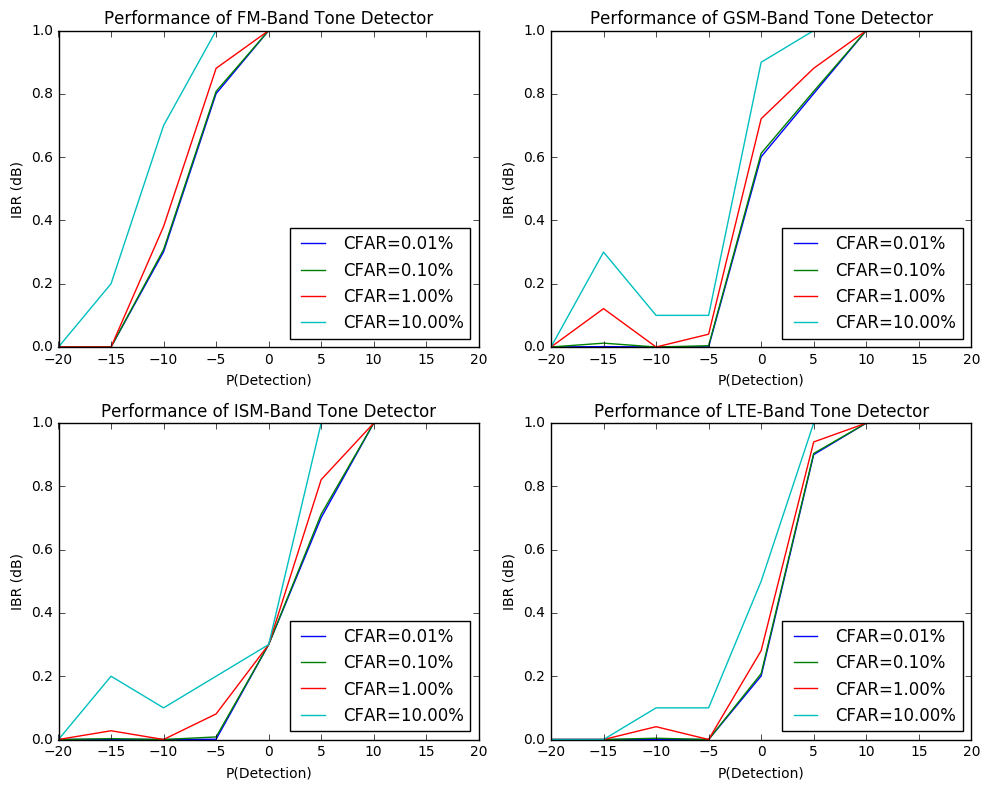}
\caption{LTE Detector Constant False Alarm Rate for LSTM Model}
\label{fig:cfarlte}
\end{figure*}

To summarize these performance behaviors into a more concise performance number, we fix a constant false alarm rate for comparison of detection performance.  In figure \ref{fig:cfarlte} we show how detection performance varies across a range of constant false alarm rates for the LTE band using the LSTM model.  

\begin{figure*}[!ht]
\centering
\includegraphics[width=0.75\textwidth]{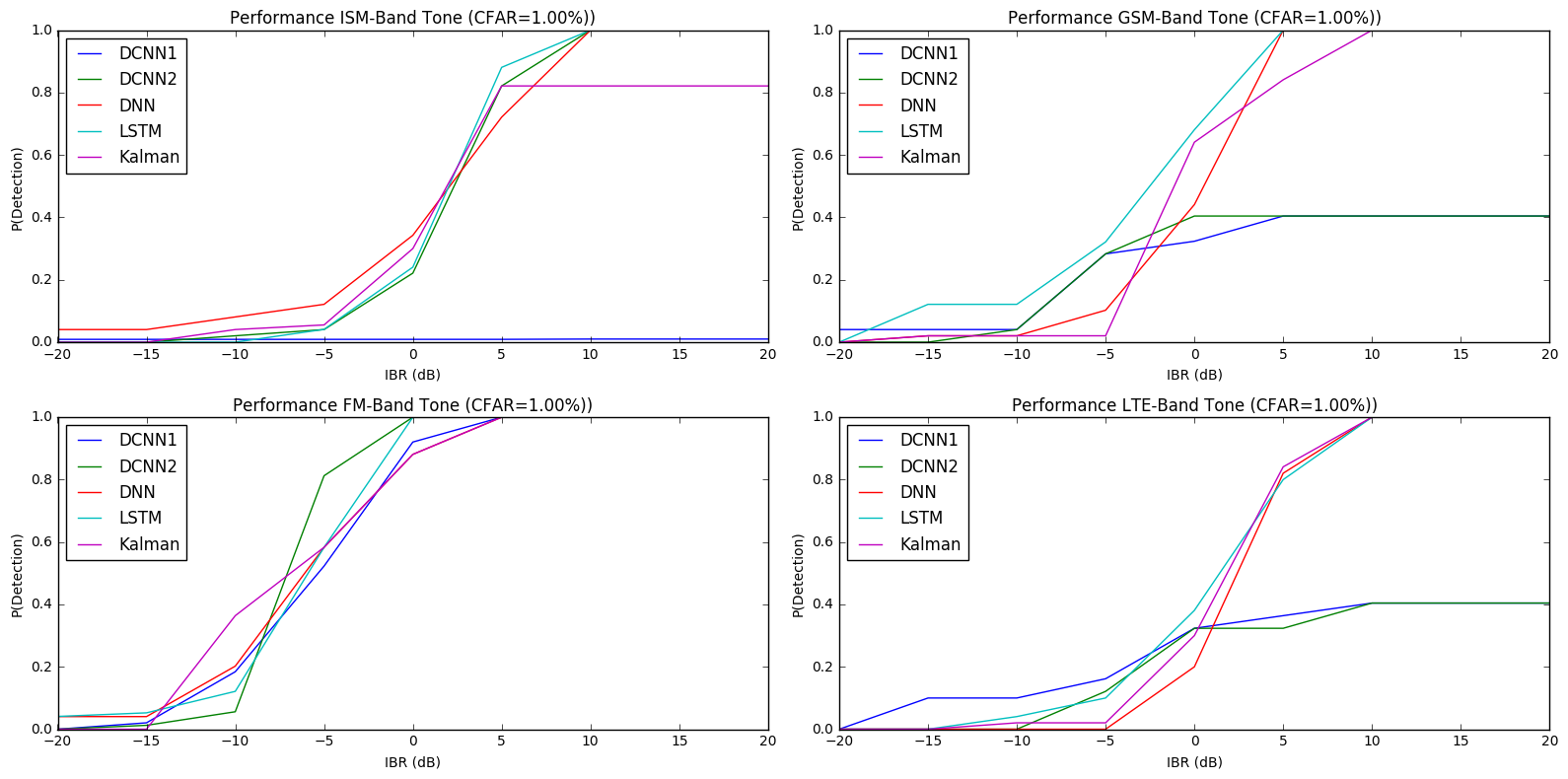}
\caption{Constant False Alarm Rate Comparison of Prediction Models}
\label{fig:cfarplot}
\end{figure*}

By repeating this for all models on all band-types, we can then pick a constant false alarm rate to compare performance across our different models.  Doing this allows us to compare model performance in different types of emitter and channel environments.  

Looking at these results, we see that in most cases the neural network based predictors outperform the Kalman based predictors slightly.  In the case of cellular networks, both GSM and LTE where a much more regular and structured temporal pattern on each carrier exists, we see slightly larger improvements in performance, likely due to having better learned a temporal predictive model suited to this behavior. 

\section{Conclusion}

In this paper we have shown how the neural network reconstruction-based anomaly detector can be used on several real wideband over the air radio bands of interest to detect anomalies occurring within band.  The results have shown that especially in structured radio signal environments where temporal sequence model prediction performs best, we obtain our best performance advantage over Kalman novelty detector methods.

We believe this is an important result that shows viability of this form of spectrum change monitoring and provides some starting points for improvements on more traditional methods for time series change detection.  We have evaluated several neural predictor models and have shown that both the LSTM model and potentially the DCNN model are viable at low SNR levels, while for an analog modulation (FM Broadcast), there was less difference between the performance of the detectors with these candidate networks.  

In future work we hope to perform much more extensive architecture and hyper-parameter searches, evaluate longer runs, larger datasets and additional types of anomalies and mixtures of anomalies.  We would like to evaluate hybrid architectures such as the LSTM with convolutional features on the front end, including both the use of dilated convolutional layers and residual units combining a number of the promising techniques which have largely been evaluated separately here.

In the area of spectrum sensing for communications system failure, interference, security, or monitoring, we hope that this method helps imagine a promising path forward towards general learning of non-signal and non-band specific methods which can be used rapidly on a wide range of systems and deployment models without needing specialized expert prior knowledge of the system of interest.

\section*{Acknowledgment}

The authors would like to thank the Bradley Department of Electrical and Computer Engineering at the Virginia Polytechnic Institute and State University, the Hume Center, and DARPA all for their generous support in this work.

This research was developed with funding from the Defense Advanced Research Projects Agency's (DARPA) MTO Office under grant HR0011-16-1-0002. The views, opinions, and/or findings expressed are those of the author and should not be interpreted as representing the official views or policies of the Department of Defense or the U.S. Government.

\printbibliography
\end{document}